\documentclass[conference]{IEEEtran}
\IEEEoverridecommandlockouts
\usepackage{apacite}
\usepackage{hyperref}
\usepackage{natbib}
\usepackage{amsmath,amssymb,amsfonts}
\usepackage{mathtools}
\usepackage{algorithmic}
\usepackage{algorithm}
\usepackage{graphicx}
\usepackage{textcomp}
\usepackage{xcolor}

\hypersetup{
  colorlinks=true,
  allcolors=blue, 
  linktoc=all
}

\renewcommand{\citeA}{\citet} 
\renewcommand{\cite}{\citep} 

\def\BibTeX{{\rm B\kern-.05em{\sc i\kern-.025em b}\kern-.08em
    T\kern-.1667em\lower.7ex\hbox{E}\kern-.125emX}}

\begin{document}

\title{Multi-gauge Hydrological Variational Data Assimilation: Regionalization Learning with Spatial Gradients using Multilayer Perceptron and Bayesian-Guided Multivariate Regression
}

\author{
\IEEEauthorblockN{
Ngo Nghi Truyen Huynh$^1$,
Pierre-André Garambois$^{1,*}$,
François Colleoni$^1$,
Benjamin Renard$^1$,
Hélène Roux$^2$
}\\
\IEEEauthorblockA{
$^1$\textit{INRAE, Aix-Marseille Université, RECOVER, 3275 Route Cézanne, 13182 Aix-en-Provence, France}\\
$^2$\textit{Institut de Mécanique des Fluides de Toulouse (IMFT), Université de Toulouse, CNRS, 31400 Toulouse, France}
}\\
\small{$^*$corresponding author: \texttt{pierre-andre.Garambois@inrae.fr}}
}

\maketitle

\begin{abstract}
  Tackling the difficult problem of estimating spatially distributed hydrological parameters, especially for floods on ungauged watercourses, this contribution presents a novel seamless regionalization technique for learning complex regional transfer functions designed for high-resolution hydrological models. The transfer functions rely on: (i) a multilayer perceptron enabling a seamless flow of gradient computation to employ machine learning optimization algorithms, or (ii) a multivariate regression mapping optimized by variational data assimilation algorithms and guided by Bayesian estimation, addressing the equifinality issue of feasible solutions. The approach involves incorporating the inferable regionalization mappings into a differentiable hydrological model and optimizing a cost function computed on multi-gauge data with accurate adjoint-based spatially distributed gradients.
\end{abstract}


\section{Introduction}

Regardless of the improvements made in hydrological forward models and available data, hydrological calibration remains a challenging ill-posed inverse problem faced with the equifinality \cite{Beven_hess_2001} of feasible solutions.
Most calibration approaches aim to estimate spatially uniform model parameters for a single gauged catchment, resulting in piecewise constant discontinuous parameters fields for adjacent catchments. Moreover, these calibrated parameter are not transferable to ungauged locations, which represents the majority of the global land surface \cite{fekete2007current, hannah2011large}. Therefore, prediction in ungauged basins remains a key challenge in hydrology \cite{Hrachowitz_2013}.

Regionalization approaches are employed to estimate hydrological model parameters in ungauged locations by transferring hydrological information from gauged locations. In early studies, the predominant method for regionalization involved individually calibrating catchments and then using multiple regression or interpolation techniques to transfer the calibrated parameter sets from gauged to ungauged locations \cite{abdulla1997development, seibert1999regionalisation, parajka2005comparison, razavi2013streamflow, parajka2013comparative}. This process can be referred to as post-regionalization \cite{samaniego2010multiscale}. However, post-regionalization approaches are limited to lumped parameters by catchment, thus ignoring within-catchment variabilities \cite{samaniego2010multiscale, razavi2013streamflow}. Furthermore, they are generally faced with the issue of equifinal parameter sets and hence equifinal estimated transfer laws, while spatial proximity is more adapted to densely gauged river networks and regions \cite{oudin2008spatial,Reichl_2009}. 
A simultaneous regionalization approach, which involves optimizing a mapping between physical descriptors and model parameters (cf. \citeA{parajka2005comparison,gotzinger2007comparison}), is able to overcome most of the aforementioned problems and can be referred as "pre-regionalization". Typically, a Multiscale Parameter Regionalization (MPR) method, combining descriptors upscaling and pre-regionalization function in form of multi-linear regressions, implemented within a spatially distributed multiscale hydrological model (mHm), has been proposed by \citeA{samaniego2010multiscale}, and later applied to other gridded hydrological models in several applicative studies (e.g., \citeA{mizukami2017towards, beck2020global}). In all the above studies, state of the art optimization algorithms are used, especially Shuffle Complex Evolution algorithm (SCE) \cite{Duan1992_SCE} in \citeA{mizukami2017towards} or Distributed Evolutionary
Algorithms (DEAP) \cite{fortin2012_DEAP} in \citeA{beck2020global}. Nevertheless, those optimization algorithms are limited to low-dimensional controls, which imposes the use of a limited number of descriptors in lumped multivariate pre-regionalization mappings, and thus restricts the capability to fully exploit the large amount of information available from multiple data sources with flexible formulations and adequate spatial rigidity. 

In \citeA{huynh2023learning}, efficient pre-regionalization algorithms have been proposed for spatially distributed hydrological modeling based on descriptors-to-parameters mappings with neural networks or multivariate regressions in a variational data assimilation framework. 
Despite the strong spatial constrain and regularizing effect introduced via pre-regionalization mappings, some sensitivity to prior remains in context of equifinality (model structural equifinality plus spatial equifinality) and its inference is explored here using the Bayesian weighting approach proposed in  \citeA{chelil2022variational,gejadze2022new}.

In this work, we present a novel seamless regionalization method for learning the pre-regionalization mapping between physical data and conceptual parameters of spatially distributed hydrological models using information from multi-gauge river flow observations and high-resolution physical descriptors. We explore two approaches to infer the pre-regionalization mapping:
\begin{itemize}
    \item Bayesian-Guided Multivariate Regional Regression (BGM2R): a multivariate polynomial regression approach, which combines high-dimensional optimization algorithms guided by a Bayesian estimation on the first guess;
    \item Artificial Neural Network Regionalization (ANNR) enabling a "seamless flow of gradient computation" and employing machine learning optimizers.
\end{itemize}

The proposed algorithms are implemented in the SMASH platform (see online documentation and tutorials at \url{https://smash.recover.inrae.fr}) available on public GitHub (\url{https://github.com/DassHydro-dev/smash}).

\section{Methodology}

\begin{figure*}[ht!]
 \centering
 \includegraphics[width=14.5cm]{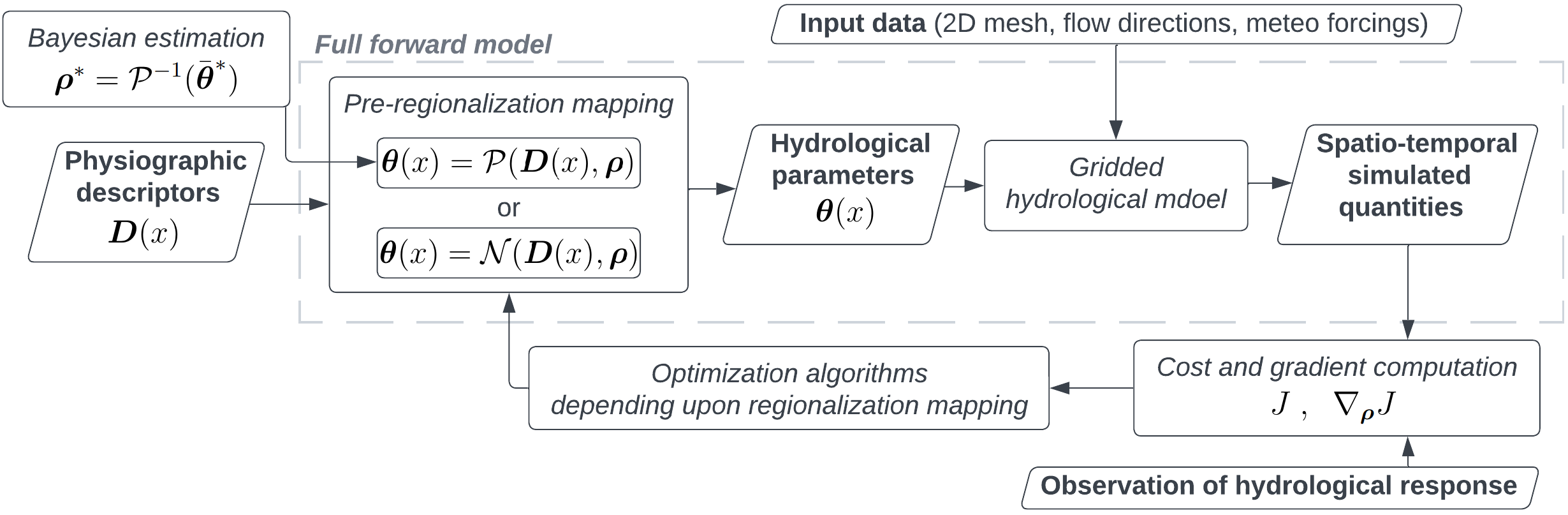}
 \caption{Flowchart of inverse algorithms for the full forward model consisting of a GR-based gridded hydrological model (spatio-temporal regular grid at 1~$\mathrm{km}^2$ and 1~$\mathrm{h}$) and a pre-regionalization mapping.}
 \label{fig:flowchart_pre-regio}
\end{figure*}

The full forward model and the optimization process are schematized in Figure \ref{fig:flowchart_pre-regio}.

\subsection{Forward Model and Cost Function}

Let us consider observed discharge time series $Q^*_g(t)$ at $N_{G}$ observation cells of coordinates $x_{g}\in \Omega$, $g\in1,..,N_{G}$ with $N_{G} \geq 1$.
For each observation cell, the corresponding gauged upstream sub-catchment is denoted $\Omega_{g}$ so that $\Omega_{ung} = \Omega \setminus \left( \cup_{g=1}^{N_{G}}\Omega_{g} \right)$ is the remaining ungauged part of the whole spatial domain $\Omega$. 
Then, the rainfall and potential evapotranspiration fields are respectively denoted as $\boldsymbol{P}\left(x,t\right)$ and $\boldsymbol{E}\left(x,t\right)$, $\forall x\in \Omega$.
The classical forward model $\mathcal{M}_{rr}$ is a dynamic operator projecting the input fields $\boldsymbol{P}\left(x,t\right)$ and $\boldsymbol{E}\left(x,t\right)$, given an input drainage plan $\mathcal{D}_{\Omega}\left(x\right)$, 
onto the discharge field $Q\left(x,t\right)$ and states fields $\boldsymbol{h}\left(x,t\right)$ written as a multivariate function:
\begin{equation}\label{eq:forward hydrological model}
\begin{aligned}
\left(\boldsymbol{h},Q\right)\left(x,t\right)=\mathcal{M}_{rr}[ & \mathcal{D}_{\Omega}\left(x\right) ,\boldsymbol{P}\left(x,t'\right),\boldsymbol{E}\left(x,t'\right), \\
& \boldsymbol{h}\left(x,0\right),\boldsymbol{\theta}\left(x\right),t], \forall (x, t') \in \Omega\times\left[0,t\right]
\end{aligned}
\end{equation}
where $\boldsymbol{\theta}$ is the $N_{\theta}$-dimensional vector of model parameters 2D fields that we aim to estimate regionally with the new algorithms proposed below, and $\boldsymbol{h}$ is the $N_{S}$-dimensional vector of internal model states. 
In this study, the distributed hydrological model $\mathcal{M}_{rr}$ is a parsimonious GR-like conceptual structure with the parameters vector 
$
    \boldsymbol{\theta}\left(x\right)= \left( c_p (x), c_{ft} (x), k_{exc} (x), l_{r} (x) \right)^T, \forall x\in \Omega
$, which is the "gr-b" structure presented in \citeA{smash2023}.

Now, the full forward model $\mathcal{M}$ is composed of the distributed hydrological model $\mathcal{M}_{rr}$ on top of which is applied a pre-regionalization operator $\mathcal{F}_{R}$ to estimate hydrological parameters $\boldsymbol{\theta}$ such that:
\begin{equation}\label{eq:forward model}
\mathbb{\mathcal{M}}=\mathcal{M}_{rr}\left[\;.\;,\;\boldsymbol{\theta}\left(x\right)=\mathcal{F}_{R}(\boldsymbol{D}\left(x\right),\boldsymbol{\rho})\right],\,\forall x \in \Omega
\end{equation}
This allows to constrain spatially and explain these spatial fields of conceptual model parameters $\boldsymbol{\theta}(x)$ from physical descriptors $\boldsymbol{D}(x)$. The pre-regionalization operator $\mathcal{F}_{R}$ being a descriptor-to-parameters mapping, with $\boldsymbol{D}$ the $N_{D}$-dimensional vector of physical descriptor maps covering $\Omega$, and $\boldsymbol{\rho}$ the vector of tunable regionalization parameters that will be defined later.

A calibration cost function is defined in order to measure the misfit between simulated and observed discharge time series, respectively denoted $Q_{g}(t)$ and $Q_{g}^{*}(t)$, for $g\in 1.. N_{G}$ gauged cells. A convex differentiable objective function is classically defined as follows:
$
    J=J_{obs}+\gamma J_{reg}
$, 
with $J_{obs}$ the observation term that measures the difference between observed and simulated quantities and $J_{reg}$ a regularization term weighted by $\gamma>0$.
The observation term is $J_{obs}=\sum_{g=1}^{N_{G}}w_{g}J_{g}^{*}$
with $w_{g}$ a physical weighting function, $J_{g}^{*}$ a local quadratic metric "at the station" (e.g., $1-NSE$) involving the response of the direct model. Thus, $J_{obs}$ depends on the control vector $\boldsymbol{\rho}$ through the direct model $\mathbb{\mathcal{M}}$.
The multi-site calibration corresponds to $N_{G}>1$ while $N_{G}=1$ is a classical calibration on a single station where $w_{1}=1$. For $N_{G}>1$, several physical weighting expressions $w_{g}$ are possible with the single constraint that $\sum_{g=1}^{N_{G}}w_{g}=1$. In this work, we simply use use $w_{g}=\frac{1}{N_{G}}$ for multiple gauges calibration.
The regional optimization problem writes as follows:
\begin{equation}\label{eq:general inv pb}
\boldsymbol{\hat{\rho}}=\arg\min_{\boldsymbol{\rho}}J\left(\boldsymbol{\rho}\right)
\end{equation}

\subsection{Regional Calibration with BGMR}

In this case, the pre-regionalization mapping $\mathcal{F}_{\mathcal{R}}\equiv\mathcal{P}$ with tunable parameter $\boldsymbol{\rho}$ consists in a multivariate polynomial regression between input physical descriptors $\boldsymbol{D}(x)$ and hydrological model parameters $\boldsymbol{\theta}(x,\boldsymbol{D},\boldsymbol{\rho}) \coloneqq \mathcal{P}\left(\boldsymbol{D}(x), \boldsymbol{\rho}\right)$ such that:
\begin{equation}\label{eq:prereg_lphp}
\begin{aligned}
       \theta_{k}(x,\boldsymbol{D},\rho_k)\coloneqq s_{k}\left(\alpha_{k,0}+\sum_{d=1}^{N_{D}}\alpha_{k,d}D_{d}^{\beta_{k,d}}(x)\right), \\
       \forall k \in[1..N_{\theta}],\forall x \in \Omega
\end{aligned}
\end{equation}
with $s_{k}(.)$ a Sigmoid-based transformation imposing bound constraints in the direct hydrological model. The lower and upper bounds are assumed to be spatially uniform for each parameter field $\theta_{k}$ of the hydrological model. 
The optimization of the control vector $ \boldsymbol{\rho} \equiv \left[\left(\rho_k\right)_{k=1}^{N_\theta}\right]^{T} \equiv \left[\left(\alpha_{k,0},\left(\alpha_{k,d}\right)_{d=1}^{N_D}\right)_{k=1}^{N_\theta}\right]^{T} $, that is solving problem \ref{eq:general inv pb}, is performed using the L-BFGS-B algorithm \cite{Zhu1997}, adapted to high-dimensional controls, without bound constraints on the $\alpha_{k,.}$, whereas the exponents $\beta_{k,d}$ is simply fixed to $1$ (multi-linear pre-regionalization). This algorithm requires the gradient of the cost function with respect to the sought parameters $\nabla_{\boldsymbol{\rho}} J $. This gradient is computed by solving the adjoint model, which is obtained by automatic differentiation using the Tapenade engine \cite{hascoet2013tapenade}. The entire process is implemented in the SMASH Fortran source code, where the full forward model $\mathcal{M}\equiv \mathcal{M}_{rr}\left(.,\mathcal{P}\left(.\right)\right)$ is a composition of both the hydrological model and the polynomial descriptors-to-parameters mapping. The convergence criterion involves reaching a maximum iteration limit or meeting conditions related to cost function change or gradient magnitude.

It is worth noting that determining a background value $\boldsymbol{\rho}^{*}$ is important for the convergence of this algorithm. It is used as a starting point for the optimization, and is defined from a spatially uniform prior $\bar{\boldsymbol{\theta}}^*$ as $\boldsymbol{\rho^*}\equiv\left[ \alpha_{k,0} = s^{-1}_{k}\left( \bar{\theta_{k}}^* \right),\left( \alpha_{k,d} = 0, \beta_{k,d} = 1\right)\right]^{T},\forall(k,d)\in[1..N_{\theta}]\times[1..N_{D}]$, where $s^{-1}_{k}(z)= \ln\left(\frac{z-l_{k}}{u_{k}-z}\right)$ is the inverse Sigmoid. 
The spatially uniform low-dimensional (LD) prior $\bar{\boldsymbol{\theta}}^*$ is determined considering the cost function without pre-regionalization, i.e., $\mathcal{M}\equiv \mathcal{M}_{rr}$ and $\boldsymbol{\rho} \coloneqq\overline{\boldsymbol{\theta}}$, and classically using a global optimization algorithm (SBS in \citeA{Michel1989}).

A Bayesian-like estimator (cf. \citeA{gejadze2022new, chelil2022variational}) is used to look at the mean of the posterior distribution $    f\left(\boldsymbol{\theta}|\boldsymbol{Q}^*\right)$ that is more stable in context of equifinality than searching its mode (inverse problem \ref{eq:general inv pb}, maximum a posteriori probability (MAP) search is the essence of variational data assimilation). 
A prior probability distribution $f_{\bar{\theta}}$ is used to generate a sample of spatially uniform parameter sets $\boldsymbol{\theta}_i, \forall i \in 1..N$ within the hypercube defined by parameters bounds $[l_k,u_k], \forall k \in 1..N_{\theta}$. The likelihood function is defined as:
$
    \mathcal{L}_i^\alpha = e^{-2^{\alpha} \left(J_i/J_{\min}-1\right)^2}
$, 
where $\alpha$ is a parameter controlling the decay rate of this function that compares the value of $J_i=J\left(\boldsymbol{\theta}_i\right)$ to $J_{min}$ the minimum value of $J_i$ over the sample of $N$ parameter sets.
The posterior ensemble mean and variance are computed as follows:
\begin{equation}
   \begin{array}{c}
\bar{\boldsymbol{\theta}}^{*,\alpha}=\frac{1}{K}{\displaystyle \sum_{i=1}^{N}\left(\mathcal{L}_{\alpha}^{i}\cdot\boldsymbol{\theta}_{i}\odot f_{\bar{\theta}}\left(\boldsymbol{\theta}_{i}\right)\right)}\\
\mathrm{Var}\left(\bar{\boldsymbol{\theta}}^{*,\alpha}\right)=\frac{1}{K}{\displaystyle \sum_{i=1}^{N}\left(\mathcal{L}_{\alpha}^{i}\cdot\left(\boldsymbol{\theta}_{i}-\bar{\boldsymbol{\theta}}^{*,\alpha}\right)\odot\left(\boldsymbol{\theta}_{i}-\bar{\boldsymbol{\theta}}^{*,\alpha}\right)\odot f_{\bar{\theta}}\left(\boldsymbol{\theta}_{i}\right)\right)}\\
\end{array}
\end{equation}
where $K=\displaystyle{\sum_{i=1}^{N} \mathcal{L}^i_\alpha \cdot f_{\bar{\theta}}\left(\boldsymbol{\theta}_i\right)}$ and $"\odot"$ denotes the Hadamard product - simple scalar product between vectors here but usable with higher dimensional controls. 
The parameter $\alpha$ is determined using the L-curve approach, considering a parametric curve $\left\{ J\left( \bar{\boldsymbol{\theta}}^{*,\alpha} \right) , D^{\alpha} \right\} $, $\alpha = -1,...,10$, where
$
    D^{\alpha}=\left(\mathrm{Var}\left(\bar{\boldsymbol{\theta}}^{*,\alpha}\right)\right)^{-1}\odot\left(\bar{\boldsymbol{\theta}}^{*,\alpha}-\bar{\boldsymbol{\theta}}^{0}\right)\odot\left(\bar{\boldsymbol{\theta}}^{*,\alpha}-\bar{\boldsymbol{\theta}}^{0}\right)
$ 
is the probabilistic (Mahalanobis) distance between the estimate $\bar{\boldsymbol{\theta}}^{*,\alpha}$ and the average prior $\bar{\boldsymbol{\theta}}^{0}=\frac{1}{N}\sum_{i=1}^{N}\boldsymbol{\theta}_i$. The value of $\alpha$ is sought in a L-curve "corner" such that it minimizes both $J\left( \bar{\boldsymbol{\theta}}^{*,\alpha} \right)$ and $D^{\alpha}$.

\subsection{Regional Calibration with ANNR}

In this case, an ANN-based regional mapping $\mathcal{F}_{\mathcal{R}}\equiv\mathcal{N}$, consisting of a multilayer perceptron, aims to learn the descriptors-to-parameters mapping such that:
   \begin{equation}\label{eq:neural_net}
       \boldsymbol{\theta}(x,\boldsymbol{D},\boldsymbol{\rho})\coloneqq \mathcal{N}\left(\boldsymbol{D}(x), \boldsymbol{W}, \boldsymbol{b}\right),\forall x \in \Omega
   \end{equation}
where $\boldsymbol{W}$ and $\boldsymbol{b}$ are respectively weights and biases of the neural network, whose output layer consists in a scaling transformation based on the Sigmoid function in order to impose bound constraints on each hydrological parameters. The regional control vector
$
   \boldsymbol{\rho} \equiv \left[\boldsymbol{W}, \boldsymbol{b}\right]^T
$
is optimized by Algorithm \ref{algo-back-prog}, that uses spatial gradients computed by the adjoint model to minimize the cost function $J(\boldsymbol{\rho})=J\big(Q^{*},\mathcal{M}_{rr}(.\; ,\; \boldsymbol{\theta}=\mathcal{N}(\boldsymbol{D},\boldsymbol{\rho}))\big)$ in the present case.
\begin{algorithm}
\caption{The proposed back-propagation at each training iteration with "opt\_func" denotes the update function of the optimizer used (e.g., Adam), and "hyper\_param" denotes its hyper parameters (e.g., learning rate)}
\label{algo-back-prog}
\begin{flushleft}
\textbf{Input:} descriptors $\boldsymbol{D}(x)$, initial/pre-updated weights and biases $\boldsymbol{\rho} = \left(\rho_1,..., \rho_{N_L}\right)$ 
\end{flushleft}
\begin{algorithmic}
  \STATE 
  $
  \boldsymbol{\theta} \gets \left[\left(\mathcal{N}\left(\boldsymbol{D}(x), \boldsymbol{\rho}\right)\right)_{x \in \Omega}\right]^T
  $
  \hfill
  $\vartriangleright$ Forward propagation
  \STATE
  $
    \nabla A \gets \nabla_{\boldsymbol{\theta}} J = \left(\frac{\partial J}{\partial \theta_1}, ...,  \frac{\partial J}{\partial \theta_{N_{\theta}}}\right)
  $
  \hfill
  $\vartriangleright$ Initialize gradient accumulation
  \FOR {$j=N_L..1$}
    \STATE 
    $
    \frac{\partial J}{\partial \rho_j} \gets \left(\frac{\partial \boldsymbol{\theta}}{\partial \rho_j}\right)^T \nabla A
    $
    \hfill
    $\vartriangleright$ Compute gradients 
    \STATE 
    $
    \nabla A \gets \nabla A . W_j^T
    $
    \hfill
    $\vartriangleright$ Update gradient accumulation
    \STATE  
    $
        \rho_j \gets \mathrm{opt\_func}\left(\frac{\partial J}{\partial \rho_j}, \rho_j, \mathrm{hyper\_param} \right)
    $ \\
    \hfill
    $\vartriangleright$ Update weights and biases
  \ENDFOR \\
\end{algorithmic}
\textbf{return} $\boldsymbol{\rho} = \left(\rho_1,..., \rho_{N_L}\right)$
\hfill
$\vartriangleright$ Updated weights and biases
\end{algorithm}

The cost function depends on the forward model $\mathcal{M}\equiv \mathcal{M}_{rr}\left(.,\mathcal{N}\left(.\right)\right)$, which is composed of two components in its numerical implementation: (i) an ANN implemented in Python, which produces the output $\boldsymbol{\theta}$ served as input for (ii) the hydrological model $\mathcal{M}_{rr}$ implemented in Fortran. To optimize $J$, we need its gradients with respect to $\boldsymbol{\rho}$. The main technical difficulty here is achieving a "seamless flow of gradients" through back-propagation. To overcome this, we divide the gradients into two parts. First, $\nabla_{\boldsymbol{\theta}} J$ can be computed via the automatic differentiation applied to the Fortran code corresponding to $\mathcal{M}_{rr}$. Then, $\nabla_{\boldsymbol{\rho}} \boldsymbol{\theta}$ is simply obtained by analytical calculus applicable given the explicit architecture of the ANN, consisting of a multilayer perceptron. The convergence criteria is simply determined by reaching the maximum number of training iterations.

\section{Results}

\subsection{Numerical experiment}

The proposed alorithms are tested on a highly challenging regionalization case from \citeA{huynh2023learning}: a high-resolution regional modeling of a flash flood prone area located in the South-East of France, with heterogeneous physical properties including karstic areas. Multiple gauges downstream of nested and independent catchments are simultaneously considered, enabling multi-gauge optimization. A total of 11 gauged catchments are employed as "donor" catchments for calibration, while 9 other catchments are treated as pseudo-ungauged for spatial validation to assess regionalization capabilities of the proposed algorithms. In this study, a set of 7 physical descriptors (see Table \ref{tab:descriptors}) available over the whole French territory is used to learn the regional transfer functions.
\begin{table}[ht!]
\caption{Physical descriptors used for pre-regionalization methods.}
\label{tab:descriptors}
\scalebox{0.9}{
\begin{tabular}{ccccc}
    \hline
    Descriptor & Type & Unit & Source \\
    \hline
     Slope & Topography & m & EU-DEM, Copernicus (2016) \\
     Drainage density & Morphology & - & \citeA{organde2013regionalisation} \\
     Karst index & Influence & \% & \citeA{caruso2013notice} \\
     Woodland percentage & Land use  & \% & CLC European Union (2012)\\
     Urbanization rate & Land use & \% & CLC European Union (2012)\\
     Soil water storage & Hydrogeology & mm & \citeA{ponce2016} \\
     Soil moisture storage & Hydrogeology & \% & \citeA{Odry2017} \\
    \hline
\end{tabular}
}
\end{table}

In the following, we compare and analyze: (i) local uniform $\boldsymbol{\rho} \equiv \bar{\boldsymbol{\theta}}$ and full spatially distributed $\boldsymbol{\rho} \equiv \boldsymbol{\theta}\left(x\right)$ calibrations for each gauges, that are respectively under- and over-parameterized hydrological optimization problems, but are served as reference performances ("Uniform (local)" and "Distributed (local)"); multigauge regional calibrations with (ii) lumped model parameters $\boldsymbol{\rho} \equiv \bar{\boldsymbol{\theta}}$ which somehow represents "level 0" regionalization ("UR"); (iii) a multivariate linear mapping (i.e., $\boldsymbol{\rho} \equiv \left[\alpha_{k, 0}, (\alpha_{k, d}, 1)\right]^T$) using a first guess obtained by global optimization algorithm ("M2R"), or guided by a Bayesian estimation ("BGM2R"); and (iv) a multilayer perceptron (i.e., $\boldsymbol{\rho} \equiv \left[\boldsymbol{W}, \boldsymbol{b}\right]^T$) ("ANNR").

Two study periods, namely P1 (August 2011 – August 2015) and P2 (August 2015 – August 2019), are considered for split sample testing. A two-fold cross-temporal calibration approach is employed, where the models are calibrated on one period and validated on the other period. In each case, we consider three types of validation: spatial validation (performance in pseudo-ungauged catchments during the calibration period), temporal validation (performance in gauged catchments during the validation period), and spatio-temporal validation (performance in pseudo-ungauged catchments during the validation period).

\subsection{Regionalization performances and analysis}

\begin{figure*}[ht!]
 \centering
 \includegraphics[width=14.5cm]{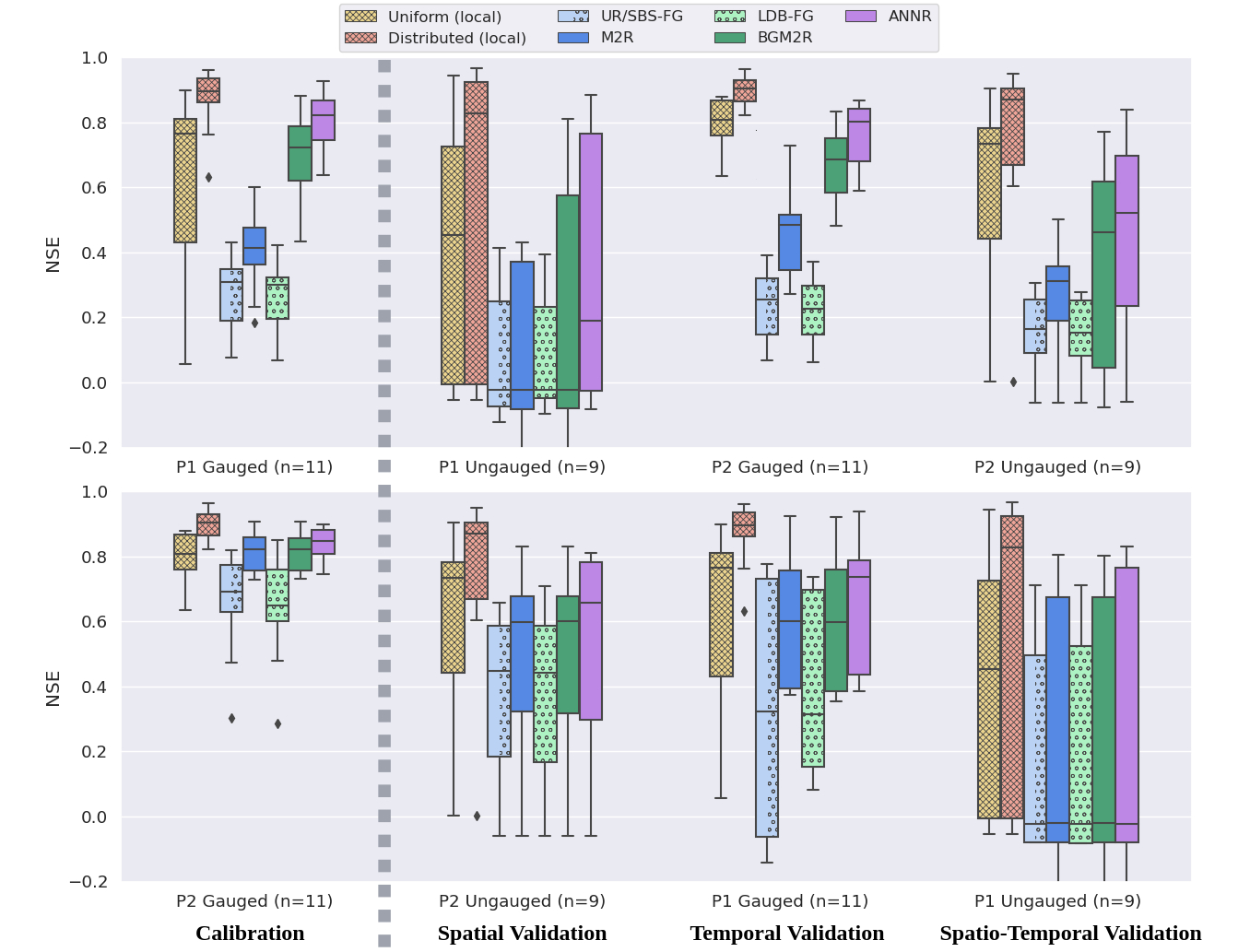}
 \caption{Performance of all calibration and regionalization methods when calibrated on P1 (upper sub-figure) and on P2 (lower sub-figure). BGM2R uses the first guess obtained by Low-Dimensional Bayesian estimation (LDB-FG), while the first guess of M2R is obtained by the SBS algorithm (SBS-FG), which also represents the solution of the regionalization method with lumped parameters (UR). $n$ denotes the number of study catchments.}
 \label{fig:boxplots}
\end{figure*}

The performance of all calibration and regionalization methods is presented in Figure \ref{fig:boxplots}. Unsurprisingly, spatially uniform calibration (UR) leads to limited performance in calibration and poor performance in regionalization, especially when compared to the reference local spatially distributed calibration that is overparameterized. The pre-regionalization methods, which incorporate information from multi-gauge discharge as well as physical descriptor maps, all yield relatively satisfying performances in calibration, temporal validation, and spatio-temporal validation at pseudo-ungauged sites (median NSE scores higher than 0.4 when calibrated on P1). The regionalization approach based on ANN (ANNR) achieves the best results for both gauged and pseudo-ungauged catchments.

Regarding the determination of prior parameter sets for the multi-linear pre-regionalization mapping, the Bayesian estimation approach (LDB-FG) demonstrates fairly good performance, comparable to that obtained with the global heuristic algorithm (SBS-FG), in calibration and spatial validation, with only minor differences in temporal validation. We believe that this is reflective of the importance of exploring a Bayesian approach for the definition of the cost function, which would enable intrinsic weighting of model misfits to different gauged hydrological behaviors. Moreover, when considering the performances of M2R and BGM2R on P1 (upper sub-figure of Figure \ref{fig:boxplots}), the Bayesian approach exhibits markedly higher performance in calibration and validations, while relatively similar performances are observed on P2 (lower sub-figure of Figure \ref{fig:boxplots}). This difference may be attributed to variations in hydrological information on P1 and potentially higher data errors, which have a lesser impact on the Bayesian approach.

\begin{figure*}[ht!]
 \centering
 \includegraphics[width=14.5cm]{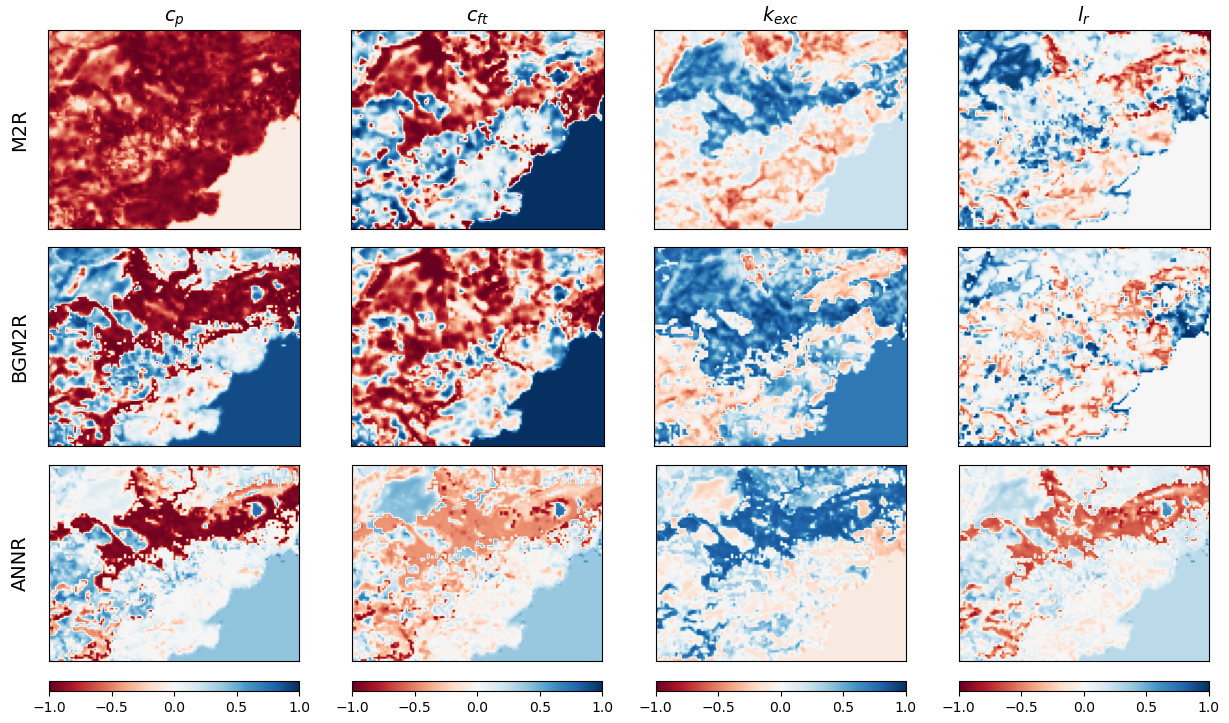}
 \caption{Normalized temporal stability over the periods P1 and P2 of the optimal distributed parameters $\frac{\hat{\theta}_k^{(P2)}-\hat{\theta}_k^{(P1)}}{u_k-l_k}, k\in [1..N_\theta]$, for the three regionalization approaches using physical descriptors.} 
 \label{fig:stab-param}
\end{figure*}

Table \ref{tab:param-space} represents several statistical quantities of the distributed parameter maps obtained through different regionalization approaches. All methods result in distinct parameter maps and varying levels of temporal stability (see Figure \ref{fig:stab-param}). The ANNR leads to the most robust inference over P1 and P2, with remarkably stable average parameter values as well as spatial standard deviation over time. The priors inferred with SBS-FG or LDB-FG exhibit slight differences and also lead to a different optimum during pre-regionalization for P1. Interestingly, the opposite trend is observed for P2, where data uncertainty and model adequacy might be better, resulting in similar functioning points after regionalization despite substantially different priors determined with SBS-FG or LDB-FG.
\begin{table*}[ht!]
\caption{The optimal parameters obtained by different methods for each calibration period. Spatially distributed parameters are represented as the median (mean, std). The rows in italic and smaller text present the spatially uniform first guess obtained by the SBS algorithm (SBS-FG) used for M2R, and the Low-Dimensional Bayesian estimation (LDB-FG) used for BGM2R.}
\label{tab:param-space}
\scalebox{1}{
\begin{tabular}{ccccc}
    \hline
    Method (Cal period) & $c_p$ & $c_{ft}$ & $k_{exc}$ & $l_r$ \\
    \hline
    UR/\textit{\scriptsize{SBS-FG}} (P1) & 2000 & 470.88 & 3.04 & 55.12 \\
    UR/\textit{\scriptsize{SBS-FG}} (P2) & 322.68 & 239.65 & -3.18 & 45.53 \\
    M2R (P1) & 2000 (2000, 0) & 113.57 (397.78, 436.27) & -0.15 (-3.8, 8.35) & 52.79 (83.63, 79.27) \\
    M2R (P2) & 398.64 (658.28, 623.15) & 338.41 (433.16, 351.08) & -0.53 (-1.56, 5.29) & 94.16 (100.66, 75.71) \\
    \textit{\scriptsize{LDB-FG}} (P1) & \textit{\scriptsize{1917.74}} & \textit{\scriptsize{527.77}} & \textit{\scriptsize{2.9}} & \textit{\scriptsize{72.32}} \\
    \textit{\scriptsize{LDB-FG}} (P2) & \textit{\scriptsize{138.53}} & \textit{\scriptsize{517.6}} & \textit{\scriptsize{-13.04}} & \textit{\scriptsize{53.71}} \\
    BGM2R (P1) & 76.02 (700.93, 869.81) & 599.93 (540.72, 404.87) & -13.33 (-10.0, 9.94) & 63.47 (92.67, 87.33) \\
    BGM2R (P2) & 396.71 (651.86, 607.05) & 348.38 (439.31, 353.62) & -0.57 (-1.47, 4.98) & 95.65 (100.98, 78.13) \\
    ANNR (P1) & 369.87 (700.95, 737.21) & 393.52 (492.71, 334.31) & -3.63 (-6.92, 7.58) & 56.63 (71.04, 62.52) \\
    ANNR (P2) & 404.13 (593.4, 477.85) & 510.9 (444.66, 220.29) & -2.23 (-2.12, 2.34) & 62.72 (67.28, 32.5) \\
    \hline
\end{tabular}
}
\end{table*}

Last but not least, during calibration on P1, M2R and BGM2R lead to a negative exchange coefficient ($k_{exc}<0$), despite starting from priors with positive exchange values ($3.04$ for SBS-FG (P1) and $2.9$ for LDB-FG (P1)); AANR also lead to negative exchange. This intriguing result, of reaching systematically significant negative exchange,  is particularly noteworthy because the exchange coefficient directly impacts mass conservation. These findings relate to those on flash floods water balance sensitivity and regionalization based on geological descriptors presented by \citeA{garambois2015parameter} for catchments in the same and nearby areas. Their event process-oriented and conservative model required an increase in modeled soil volume, while pedological and geological descriptors provided valuable constraining information, especially in the context of flash floods. 

\section{Conclusion}

A Bayesian calibration algorithm has been tested in this study, on top of our Hybrid Variational Data Assimilation Parameter Regionalization (HVDA-PR) approach enabling seamless regionalization in hydrology.
The methods were tested in a challenging flash flood-prone area in the South-East of France, characterized by diverse physical properties and hydrological responses. Overall, the methods demonstrated satisfactory performance in several aspects: (i) accurate modeling of discharge at both gauged and pseudo-ungauged sites, and (ii) effective identification of conceptual parameters and extraction of information from physical descriptors.
Notably, the ANN-based regionalization method outperformed other approaches in terms of discharge accuracy and parameter stability. Bayesian prior estimation exhibited good performance and relative robustness, even in challenging cases like calibration on the period P1, where data uncertainty and model inadequacy were assumed to be higher. While the Bayesian method is computationally more demanding than traditional low-dimensional calibration algorithms, it can be efficiently parallelized. Moreover, the Bayesian approach can be extended to higher-dimensional contexts, such as determining semi-distributed priors and exploring spatial equifinality using our variational data assimilation algorithms.
Interestingly, in contrast to the ANN, the regression methods provided insights into more complex modeling situations and potential data-model discrepancies. This highlights the importance of maintaining both "classical" approaches and AI-based solutions in research and applications, particularly in the continuous development of physically and mathematically interpretable methodologies.

\section*{Acknowledgments}
The authors greatly acknowledge SCHAPI-DGPR and Météo-France for providing data used in this work; 
Igor Gejadze for scientific discussion; SCHAPI-DGPR, ANR grant ANR-21-CE04-0021-01 (MUFFINS project, "MUltiscale Flood Forecasting with INnovating Solutions"), and NEPTUNE European project DG-ECO for funding support.

\bibliographystyle{apacite}
\bibliography{refs}

\end{document}